\documentclass[sigconf]{acmart}

\usepackage{subcaption}
\usepackage{enumitem}
\usepackage{pgfplots}
\usepackage{stfloats}
\usepackage{tcolorbox}
\pgfplotsset{compat=1.18}
\definecolor{color1}{RGB}{79, 129, 189} 
\definecolor{color2}{RGB}{155, 187, 89}  
\definecolor{color3}{RGB}{192, 80, 77}   

\AtBeginDocument{%
  }

\copyrightyear{2026}
\setcopyright{none}
\acmYear{2026}
\acmDOI{}

\acmConference[KDD '26]{The 32nd ACM SIGKDD Conference on Knowledge Discovery and Data Mining}{August 9--13, 2026}{Jeju, Republic of Korea}

\begin{document}

\title{FT-RAG: A Fine-grained Retrieval-Augmented Generation Framework for Complex Table Reasoning}


\author{Zebin Guo}
\affiliation{%
  \institution{Georgia Institute of Technology}
  \city{Atlanta}
  \state{Georgia}
  \country{USA}
}
\email{zbguo@gatech.edu} 

\author{Weidong Geng}
\affiliation{%
  \institution{Zhejiang University}
  \city{Hangzhou}
  \state{Zhejiang}
  \country{China}
}
\email{gengwd@zju.edu.cn} 

\author{Ruichen Mao}
\affiliation{%
  \institution{Zhejiang Lab}
  \city{Hangzhou}
  \state{Zhejiang}
  \country{China}
}
\email{maoruichen@zhejianglab.org} 

\begin{abstract}
Retrieval-Augmented Generation (RAG) enhances Large Language Models (LLMs) by grounding responses in external knowledge during inference. However, conventiona RAG systems under-perform on structured tabular data, largely due to coarse retrieval granularity and insufficient table semantic comprehension. To address these limitations, we introduce FT-RAG, a fine-grained framework that employs knowledge association by decomposing tables into entry-level semantic units to construct a structured graph. FT-RAG employs a structural neighbor expansion mechanism to find semantically connected entities during graph retrieval, followed by multi-modal fusion to consolidate the context of table retrieval results. Further, to address the scarcity of specialized datasets in this domain, we introduce Multi-Table-RAG-Lib, a benchmark comprising 9870 QA pairs with high complexity and difficulty, curated to demand multi-table integration and text-table information fusion for reasoning. FT-RAG surpasses top-performing baselines across all metrics, achieving a 23.5\% and 59.2\% improvement in table-level and cell-level Hit Rates, respectively. Generation performance also sees a remarkable 62.2\% increase in exact value accuracy recall. These metrics verify the framework's effectiveness in factual grounding across both pure tabular and heterogeneous table-text contexts. Therefore, our method establishes a new state-of-the-art performance for complex reasoning over mixed-modality documents.
\end{abstract}



\maketitle
\section{Introduction}
In real-world scenarios such as financial analysis and scientific review, high-value knowledge is often densely encoded in tables. For human analysts, extracting insights from these complex documents is an arduous task, requiring them to locate precise data points while simultaneously grasping their semantic links to surrounding text and related tables. To alleviate this burden, Large Language Models (LLMs) have emerged as the primary engine for automated question answering, enabling users to resolve intricate queries through natural language. However, the success of this paradigm hinges on the model’s ability to achieve a comprehensive and accurate understanding of the underlying data. Since LLMs inherently suffer from static knowledge and hallucinations \cite{arslan-2024}, Retrieval-Augmented Generation (RAG) is widely used to anchor model outputs in external evidence, ensuring that responses are factually grounded \cite{fan-2024}.

However, while typical RAG pipelines succeed in handling unstructured text, they face significant challenges when applied to the structural complexity of tables. Unlike plain text, tables encode information through multi-dimensional layouts where spatial proximity implies semantic relationships. To address this, recent research has attempted to adapt RAG for tabular data \cite{guo-2025, yu-2025}. Unfortunately, these existing table-oriented RAG methods mostly adopt a coarse-grained strategy, typically treating a table as a single unit \cite{zou-2025} or flatten it into plain text descriptions \cite{zhang-2025}, turning a structural problem into a text-based task. However, this simplification obscures the precise, cell-level relationships needed for high-accuracy reasoning. As a result, these approaches have limitations in supporting rigorous decomposition and alignment required to build strong links between tables and their surrounding text. When a query requires integrating fragmented data from heterogeneous documents, existing paradigms lack the granularity to perform the necessary cross-table information fusion and complex reasoning.

To address this gap, we propose FT-RAG (\textbf{F}ine-grained \textbf{T}able-aware \textbf{R}etrieval-\textbf{A}ugmented \textbf{G}eneration), a novel framework designed to process tabular data at finer level of details. FT-RAG uses the following key components to improve retrieval and reasoning performance:

\begin{itemize}
    \item \textbf{Entry-level table decomposition}.  FT-RAG fractures each table into cell-level \textit{semantic units} called cell groups. Each cell group encapsulates not only the cell value but also its contextual metadata.
    \item \textbf{Structured SAT graph construction}. All semantic units are organized into a unified, schema-aware \textbf{SAT-Graph}. In this hierarchical representation, internal nodes represent semantic concepts, while leaf nodes store grounded values along with their complete metadata traces.
    \item     \textbf{Joint text-table fusion for generation}. FT-RAG reconstructs textual context from original documents by associating each unit with its surrounding textual environment. This step resolves the potential ambiguity of table units. 

\end{itemize}

Beyond proposing a novel RAG framework, we also address a critical gap in benchmark. Existing public table-centric benchmarks\cite{wu2025tablebenchcomprehensivecomplexbenchmark} primarily focus on single-table retrieval, where questions typically ask for a specific atomic value and expect short, factual answers. The answers are typically designed to be compatible with simple evaluation metrics such as Exact Match. In contrast, our task requires multi-table fusion and associative retrieval: answering a query often depends on retrieving multiple interrelated evidence units from different tables, which must then be synthesized by the LLM to perform reasoning. To support this more realistic and challenging setting, we introduce Multi-Table-RAG-Lib, a new benchmark specifically designed for complex, multi-hop table QA that emphasizes cross-table relationships, contextual grounding, and compositional answer generation. Our benchmark endorse queries that are not only focused on specific cells but also to locate relevant answers and reorganize scattered values into coherent answers,providing a more rigorous and realistic testbed for evaluating the performance of RAG systems in tabular-centric environments.

\section{Related Work}
In this section, we review two lines of research related to our work: Graph-Enhanced Retrieval-Augmented Generation and Tabular LLMs.

\subsection{Graph-Enhanced Retrieval-Augmented Generation} Traditional Retrieval-Augmented Generation systems rely on dense vector similarity\cite{lewis2021retrievalaugmentedgenerationknowledgeintensivenlp} or keyword search \cite{10.1561/1500000019} to retrieve relevant documents for Large Language Models. However, standard RAG struggles with complex questions that necessitate multi-hop reasoning\cite{tang2024multihopragbenchmarkingretrievalaugmentedgeneration} to synthesize fragmented information. Graph-enhanced RAG addresses these limitations by incorporating structural priors into the retrieval process, constructing topological representations of knowledge—typically by extracting entities and relationships to form knowledge graphs\cite{zhu2025knowledgegraphguidedretrievalaugmented}. Methodologies in this domain generally aim to optimize retrieval efficiency or aggregate knowledge via multi-hop reasoning. Recent approaches range from LightRAG\cite{guo2025lightragsimplefastretrievalaugmented}, which utilizes index-based key-value search to bypass costly community detection, to HippoRAG\cite{gutierrez2025hipporagneurobiologicallyinspiredlongterm}, which leverages human memory theory for graph retrieval, and LinearRAG\cite{zhuang2025linearraglineargraphretrieval}, which employs query dependency graphs. Complementary strategies treat the graph as a navigation map to enable multi-level reasoning, as seen in RAPTOR\cite{sarthi2024raptorrecursiveabstractiveprocessing} and GraphSearch\cite{liu2026graphsearchagenticsearchaugmentedreasoning}. Despite these advances in text-centric retrieval, the scarcity of Table-centric RAG with graph augmentation restricts the exploitation of structured data, particularly given the inherent structural alignment between tabular inputs and graph models.

\subsection{Tabular LLMs} Current approaches for equipping LLMs with tabular reasoning capabilities primarily follow a \textit{Table-to-Text} paradigm, which can be categorized based on the data source: Database Tables and Document Tables. For structured databases, methods focus on serializing rigid schemas and data into natural language. For instance, RoT \cite{zhang2025rotenhancingtablereasoning} utilizes individual rows as the fundamental retrieval unit to align with vector search, while GRIT \cite{kang-etal-2025-grit} transforms relational schemas into textual representations to ground the generation process. However, these approaches are inherently constrained by strict table definitions. Conversely, document-embedded tables exhibit greater structural variability and require integration with surrounding text for context. To address this, HRoT \cite{luo2023hrothybridpromptstrategy} employs a hybrid prompting strategy combined with multi-hop reasoning to convert tabular data into coherent textual descriptions. Similarly, MixRAG \cite{zhang-2025} integrates cell's position into cell-level textual summaries to assist in generating comprehensive summaries. Despite effectively leveraging the semantic power of LLMs, these work inevitably abstracts away critical table meta-information, limiting their effectiveness in complex, multi-table reasoning scenarios.

\section{Preliminary}
In this section, we propose formal definitions of our Table QA Question Corpus, Document Corpus for retrieval, and the Table-centric RAG task. Further, we define our dataset to evaluate proposed  methodology and other baseline methods that solve the task defined.

\subsection{Definitions}
We formally define the Fine-grained Table-aware RAG task, which extends conventional table-level or chunk-level retrieval by focusing on table entry-level within complex documents containing both text and tables.

\textbf{Question Corpus.}  
Let $\mathcal{Q} = \{Q_1, Q_2, \ldots\}$ denote a corpus of user questions, where each question $Q_i = \langle q_i, f_i \rangle$ consists of a natural language query $q_i$ and a contextual flag $f_i \in \{0,1\}$. The contextual flag $f_i$ determines the evidence scope required for answering: $f_i=0$ indicates that the necessary information can be fully extracted from tabular data alone, while $f_i=1$ requires supplementary semantic information from associated textual passages in conjunction with tabular values.

\textbf{Document Corpus.} 
Let $\mathcal{C} = \{D_1, D_2, \dots, D_{|\mathcal{C}|}\}$ be a corpus of heterogeneous documents. Each document $D_i$ comprises:
\begin{itemize}
    \item A set of text segments $\mathcal{P}_i = \{P_1, P_2, \dots\}$, and
    \item A collection of tables $\mathcal{T}_i = \{T_1, T_2, \dots\}$, where each table belongs to one of two structural categories:
    \begin{itemize}
        \item \textit{Flat tables}: (a) one-dimensional key-value lists, or (b) two-dimensional relational tables with flat row/column headers;
        \item \textit{Hierarchical tables}: (c) one-dimensional tables with nested sections or grouped rows, or (d) two-dimensional tables with multi-level headers or merged cells that encode semantic hierarchy.
    \end{itemize}

\end{itemize}

\textbf{Task.}  
Given a question $Q = \langle q, f \rangle$ and a heterogeneous context corpus $\mathcal{C} = \{\mathcal{T}, \mathcal{P}\}$, where $\mathcal{T}$ denotes structured tables and $\mathcal{P}$ denotes associated textual passages, the goal is to:
\begin{enumerate}
    \item Retrieve the most relevant fine-grained evidence units—namely, (i) the table entries $e^*$ that answer $q$, and (ii) the associated textual context $p^* \in \mathcal{P}$ that clarifies the semantics of $e^*$ when the contextual flag $f = 1$;
    \item Generate a faithful answer $A$ grounded in this combined evidence.
\end{enumerate}

Formally, the task is defined as:
$$
A = F(Q, \mathcal{C}) = \textsc{Inference}(Q, e^*, p^*)$$
$$\text{where} \quad (e^*, p^*) = \textsc{Retrieve}(Q, \mathcal{C}).
$$

Here, $\textsc{Retrieve}(Q, \mathcal{C})$ identifies the minimal yet sufficient evidence span across tables and, if $f=1$, complementary text. Each retrieved table entry $e^*$ is a structured tuple that includes:
\[
e^* = (v, c, e, t, s, k, m),
\]
where $v$ is the cell value, $c$ is its column context, $e$ denotes the entity, $t$ encodes temporal information, $s$ represents the subject, $k$ is the key attribute, and $m$ is metadata used to link the entry to related textual passages in $\mathcal{P}$. When the contextual flag $f = 1$, the retriever uses $m$ to fetch the corresponding passage $p^* \in \mathcal{P}$ that provides semantic or procedural clarification.

Finally, in \textsc{Inference} phase, the question $q$, the structured evidence $e^*$, and (if applicable) the textual context $p^*$ are formatted into a unified prompt and fed to LLMs to generate the final response $A$.

\subsection{Dataset}
To the best of our knowledge, most existing datasets tailored for table-centric tasks focus on retrieving a single table value, yielding short and concise answers as ground truth. This setting is ill-suited for evaluating Table QA questions that require reasoning over cell values spanning multiple tables. To address this gap, we introduce our own dataset, \textsc{Multi-Table-RAG-Lib}. 

This dataset is derived from \textsc{MultiHiertt}~\cite{zhao-etal-2022-multihiertt}, a corpus containing 7,830 paragraphs with an average of 30.79 text chunks and 3.89 tables. The dataset satisfies our structural requirements by featuring naturally interleaved tables—both flat and hierarchical—within textual passages, fostering deep semantic interdependency. To address the issue of missing subject entities in the original anonymized data, we utilize an LLM to infer and restore the central topic from the document context, ensuring semantic integrity without polluting the dataset with external hallucinations. Furthermore, to strictly enforce a table-centric evaluation and mitigate information redundancy, we systematically filter the textual context. Specifically, we re-paraphrase textual sentences that contain exact numerical values present in the tables using LLMs. This pre-processing step ensures that the model is compelled to derive precise quantitative data exclusively from the tabular structures, rather than relying on textual mentions, thereby isolating the model's tabular reasoning capabilities. Details are revealed in Appendix \ref{app:data_prep}

To bridge questions corpus, we select different table description fields, accompanied with original HTML5 table representation of \textsc{MultiHiertt} to construct questions and corresponding ground truth answers. Instead of random selection, we carefully find common characteristics of different fields, and align the selection by fixing some features. We select random fields that share the same date, with the same subject, or the same entity. The degree of selected fields and random pairs are set by multiple hyperparameter combinations, in order to reflect different degrees of question difficulty. We generate 4935 QA pairs. These pairs' information is solely from table cells, in order to reflect the performance of pure table retrieval in multi-table RAG tasks, i.e. the contextual flag $f$ is set to 0. Further, we enhance these QA pairs by providing surrounding text of applied tables, adding semantic context in text to the questions, to evaluate the performance of our task in table-text combined corpus, i,e. the contextual flag $f$ is set to 1. The total number of valid QA pairs is 9870, all questions are generated from 5000 table fields within 308 large tables across 81 documents. Detailed statistics can be found in Appendix \ref{app:dataset_stats}
\begin{figure}
    \centering
    \includegraphics[width=1\linewidth]{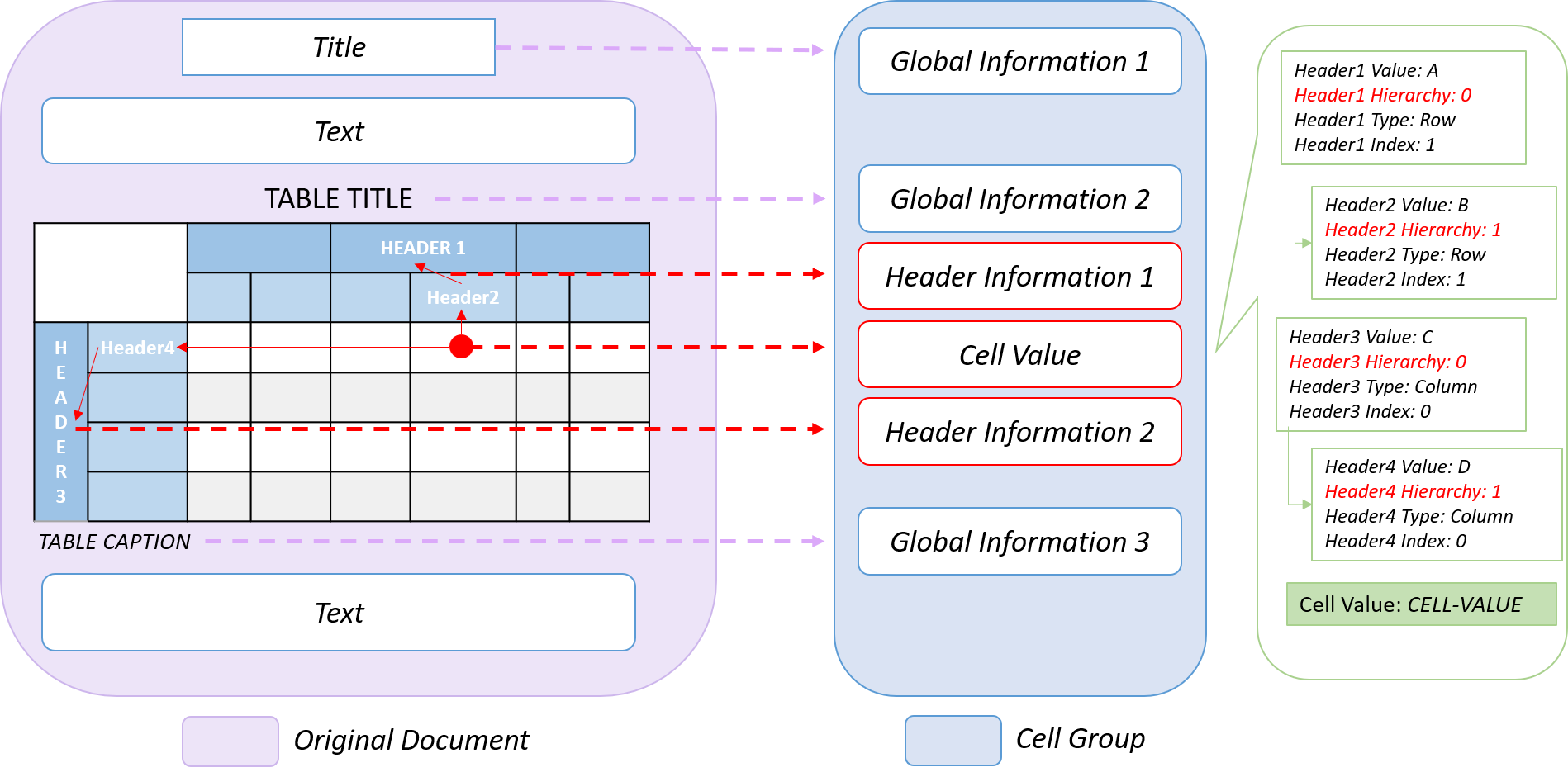}
    \caption{\small \textbf{Fine-grained table parsing mechanism.} The framework transforms individual cells into metadata-enriched \textbf{cell groups}. The associated header information explicitly encodes \textbf{indices}, \textbf{types}, and \textbf{hierarchical tiers} to ensure precise structural grounding and semantic alignment.}
    \label{fig:parsing_mechanism}
\end{figure}
 
\section{Methodology}

To address this challenging setting, we propose FT-RAG, a novel framework tailored for table-centric RAG task. FT-RAG is specially designed to handle heterogeneous corpora comprising interlinked tables and text. To answer queries under this setting, FT-RAG is capable of identifying semantically complete evidence units at the cell level across multiple heterogeneous tables, and dynamically incorporating complementary textual context when required for disambiguation or semantic enrichment. FT-RAG consists of three core modules:
\begin{itemize}
    \item  A \textbf{Fine-Grained Table Parsing Module} that decomposes raw tables into structured semantic units;
    \item  A \textbf{Structure-Aware Semantic Lifting Module} that organizes these units into a schema-aware temporal and knowledge graph grounded in analytical dimensions;
    \item  A \textbf{Graph Retrieval and Multi-Modal Fusion Module} that first performs layered search over SAT graph to retrieve atomic facts, then fuses retrieved tabular evidence with relevant textual context
\end{itemize}

\subsection{Fine-Grained Table Parsing Module}

We manually reviewed thousands of official documents, including financial statements, research reports and academic papers, to propose a sementic atomization pipeline that decompose raw tables in documents into  metadata-enriched units.

 The decomposition process begins by converting PDF documents into machine-friendly Markdown formats using high-fidelity tools such as MinerU~\cite{wang2024mineruopensourcesolutionprecise}. We first extract global metadata, including document titles, table captions, and surrounding context, to generate a schema applicable to every cell in a table.

Subsequently, we implement entry-level metadata anchoring. Rather than processing the table as a monolithic block, we decompose it into independent units. Specifically, we transform each individual cell into a Cell Group, which encapsulates the raw cell value with its full context—including global document metadata, table-level metadata, and hierarchical header paths.

As illustrated in Figure~\ref{fig:parsing_mechanism}, to preserve the table's structural logic, the associated header information explicitly encodes three key attributes:
\begin{itemize}[nosep]
    \item \textbf{Indices}: The absolute coordinates originating from the top-left anchor.
    \item \textbf{Types}: The semantic role (e.g., row header vs. column header).
    \item \textbf{Hierarchies}: The hierarchical tier depth, indicating the nesting level.
\end{itemize}
This storage ensures every cell carries sufficient context for downstream graph construction, which decomposes the raw textual and structural segments into atomic semantic primitives.

\subsection{Structure-Aware Semantic Lifting Module}
\begin{figure*}
    \centering
    \includegraphics[width=0.9\linewidth]{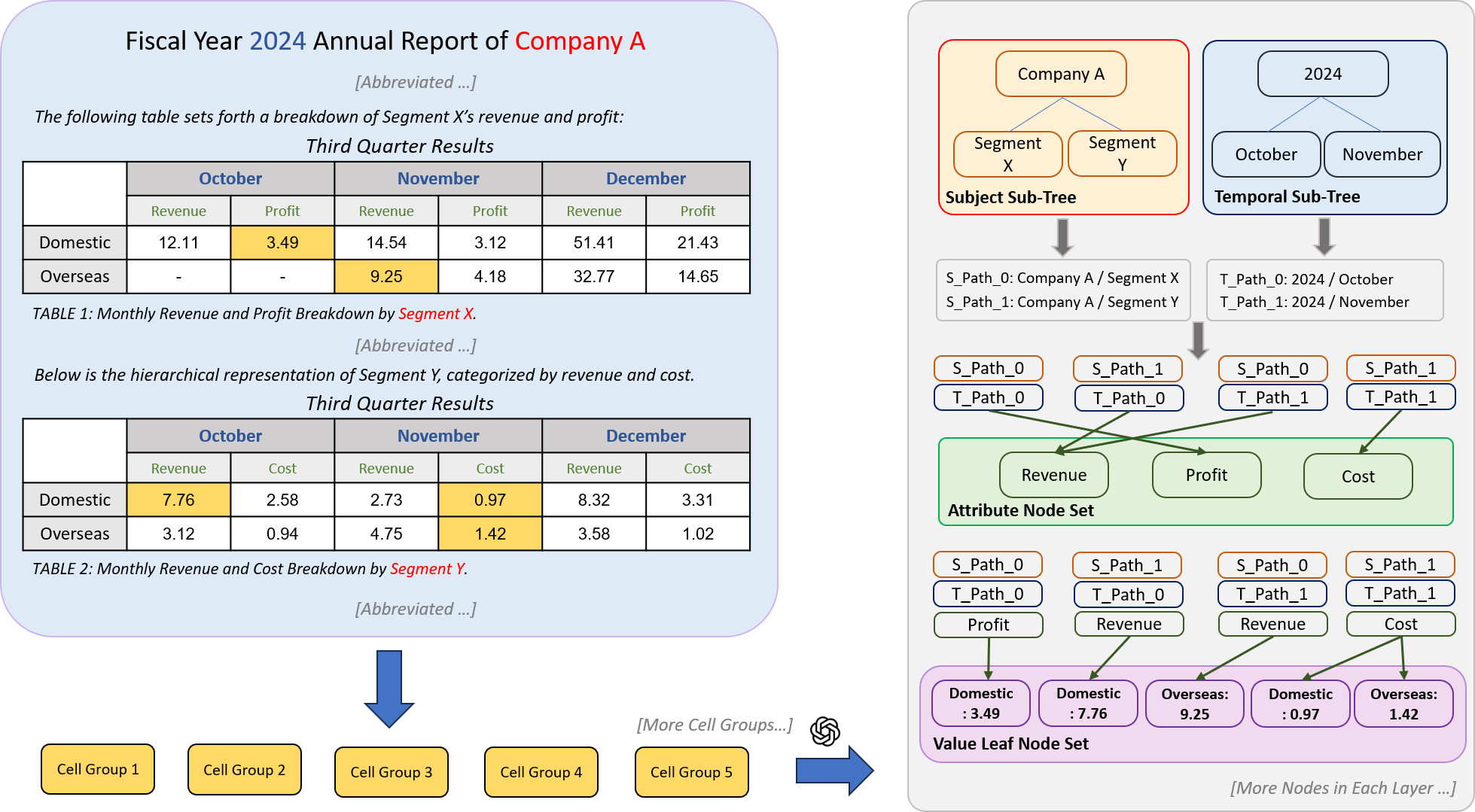}
    \caption{Schematic illustration of the \textbf{Semantic Lifting Mechanism}. The framework constructs a SAT Graph by decomposing cell groups into four distinct components: the Subject Sub-tree ($\mathcal{T}_S$, yellow) generated via LLM inference for structural lineage; the Temporal Sub-tree ($\mathcal{T}_T$, blue) derived from pattern-based parsing; the pivotal Attribute Node Set ($\mathcal{V}_A$, green) serving as semantic anchors; and the atomic Value Leaf Node Set ($\mathcal{V}_L$, pink) attached via composite indices.}
\label{fig:semantic_lifting}
\end{figure*}
Existing table understanding methods typically adopt linearization strategies to flat 2-D tabular data into 1-D textual descriptions, we argue that this approach will lead to information loss, such as the orthogonal relationship between row and column header attributes. Furthermore, as emphasized in Temporal Augmented Retrieval \cite{zhu2025rightanswerrighttime}, traditional RAG frameworks operate under a timeless assumption, neglecting the critical dimension of temporal validity. This oversight inevitably degrades retrieval performance, particularly when targeting table entries, as tabular data often represents high-precision snapshots of dynamic metrics. A numerical entry is only factually accurate within its specific time window.

To tackle these limitations, we propose a novel Semantic Lifting Mechanism to transform discrete cell groups into interconnected knowledge units. Our approach is grounded in three critical dimensions necessary for data reconstruction: \textbf{S}ubject, \textbf{A}ttribute, and \textbf{T}emporal. By explicitly anchoring each cell to these dimensions, we prevent context drift common in purely textual retrieval.

Formally, we define the SAT (Subject-Attribute-Temporal) Graph as a composite directed acyclic graph, denoted as $\mathcal{G} = (\mathcal{V}, \mathcal{E})$. The topology is constructed by anchoring two hierarchical sub-trees, the Subject Sub-tree $\mathcal{T}_S$ and the Temporal Sub-tree $\mathcal{T}_T$, onto a pivotal Attribute Vertex Set $\mathcal{V}_A$. The Value Leaves $\mathcal{V}_L$ are attached to the SAT graph. The graph construction follows a convergent decomposition strategy:
\begin{itemize}
    \item \textbf{Subject Sub-tree ($\mathcal{T}_S$)}: Corresponds to the Subject dimension. This structure is instantiated via LLM inference over the Cell Group's global context. Crucially, the LLM not only extracts individual entity identities but also reconstructs their inherent structural lineage. For instance, it establishes hierarchical dependencies where "Laptop" is nested under "Hardware".
    \item \textbf{Temporal Sub-tree ($\mathcal{T}_T$)}: Corresponds to the Temporal dimension. This structure organizes discrete time intervals or validity scopes chronologically for time-series navigation. Temporal values are generated via pattern-based parsing and normalization, creating an inherently nested representation. For example, '2025-01-10' is positioned as a specific instance within the '2025' parent interval.
    \item \textbf{Attribute Node Set ($\mathcal{V}_A$)}: Corresponds to the Attribute dimension. Nodes in this set, $a_k \in \mathcal{V}_A$, act as semantic convergence points for the graph. Formally, an attribute node is defined as the anchor for specific paths originating from the sub-trees, denoted as a mapping $\phi: (s_i, t_j) \to a_k$, where $s_i \in \mathcal{T}_S$ and $t_j \in \mathcal{T}_T$. This mechanism allows a single attribute $a_k$ to bridge disparate subject-temporal contexts, effectively fuse fragmented entities and timelines for cross-contextual retrieval.
    \item \textbf{Value Leaf Node Set ($\mathcal{V}_L$)}: Stores the atomic data entries as key-value pairs. Rather than existing as isolated vertices, these nodes are attached to specific SAT combinations. We denote this relationship as a surjective mapping $\mathcal{H}: \mathcal{T}_S \times \mathcal{T}_T \times \mathcal{V}_A \to \{v_1, v_2, \dots\}$, where a unique path of subject, temporal, and attribute coordinates (i.e., $\langle s_i, t_j, a_k \rangle$) serves as the composite index that anchors a set of value leaves. This structure ensures that every data point is retrievable via its semantic address. 
\end{itemize}

More concretely, the metadata enriched in cell groups serve as the structural foundation for semantic lifting. Specifically, the \textbf{Types} attribute acts as a semantic heuristic, guiding the LLM to accurately distinguish between temporal markers  and categorical attributes. Simultaneously, the \textbf{Indices} and \textbf{Hierarchies} serve as a structural blueprint that, when coupled with semantic inference, guides the generation of parent-child relationships. 

Building upon the local decomposition of individual cell groups, we synthesize these units into a global graph structure. As illustrated in Figure~\ref{fig:semantic_lifting}, the components outlined above are organized into three hierarchical layers:

\begin{itemize}[nosep]
    \item \textbf{The Anchor Layer}: This layer contains Subject and Temporal subtrees, designed to enforce isolation in a coarse retrieval stage.
    \item \textbf{The Semantic Intersection Layer}: Functions as a semantic buffer, bridging the gap between monolithic schema attributes and specific contextual anchors. This design prevents a meaningless enumeration of all potential fields; instead, it builds explicit links that map only relevant subjects to their instantiated values.
    \item \textbf{The Value Leaves Layer}: This terminal layer stores atomic data units as key-value pairs.
\end{itemize}

This structure facilitates retriever navigates from high-level Anchors to specific Value Leaves, treating the search process as a logical traversal over the graph. This allows for the precise isolation of specific cell groups without losing their inherent hierarchical context. Crucially, as each leaf node is bi-jective to a table cell, each leaf retains a deterministic provenance link to its source table. This design guarantees traceability in the retrieval stage, supporting robust table-text fusion in downstream stage.

\subsection{Graph Retrieval and Multi-Modal Fusion Module}

Linearization approaches often collapse structured indices into vector-similarity-based retrieval, which is ill-suited for exact value location. For instance, a marginal drift in vector space—such as conflating '2023' with '2024'—can lead to factual errors that textual relevance alone cannot correct. To address this, the retrieval process executes a structure-aware navigation over the SAT Graph. 

However, precise structural localization is not equivalent to comprehensive semantic understanding. While tabular schemas offer a compact representation, their inherent brevity often comes at the cost of informational completeness. Tables prioritize display efficiency, frequently omitting comprehensive background constraints. We argue that for an LLM to accurately interpret tabular values, the table alone is insufficient; the surrounding textual context is indispensable for resolving ambiguities. Consequently, we augment the retrieved graph nodes with their original textual narratives. This multi-modal fusion ensures that the rigorous logic of the graph is complemented by the descriptive power of text, providing the LLM with the necessary context to decode the true meaning of the retrieved values.

\begin{figure*}
    \centering
    \includegraphics[width=0.85\linewidth]{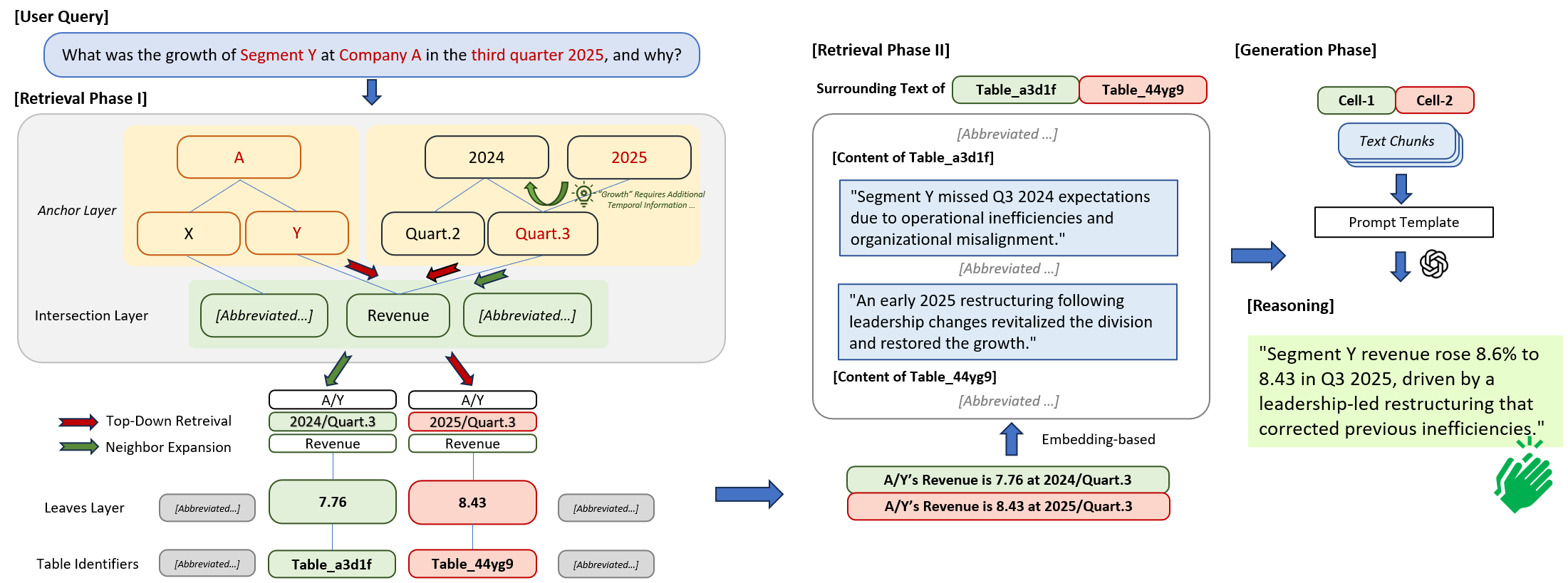}
    \caption{\textbf{Two-Phase Retrieval Mechanism.} The figure details the transition from hierarchical graph navigation to text-bridged augmentation. Structured units identified within the SAT Graph are fused with surrounding text to form a unified context }
    \label{fig:retreival_generation}
\end{figure*}

Formally, to address the inherent complexity of tabular data and the linguistic ambiguity of user queries, we propose a comprehensive retrieval paradigm comprising two distinct phases: \textbf{Hierarchical Graph Navigation} and \textbf{Text-Bridged Augmentation}. The former executes a topology-aware search to isolate valid subgraph paths and implicit semantic neighborhoods, while the latter fuses these structured facts with unstructured narratives to ground the final generation, as shown in Figure \ref{fig:retreival_generation}.
\subsubsection{Phase I: Hierarchical Graph Navigation}This phase functions as a mapping mechanism to align the user's query $q$ with the graph's dimensions: Subject ($s$), Temporal ($t$), and Attribute ($a$). To address varying degrees of query specificity—from fully constrained requests to partial references—we employ a Dual-Path Traversal Strategy that deduces missing variables through two complementary mechanisms:\textbf{(1) Anchor-Driven Forward Traversal.} This strategy is invoked when the query provides high-level contexts (Subject and Time) but lacks specific attribute definitions (e.g., \textit{"Status of Company A in Q2"}). The system initiates a top-down search from the identified $s$ and $t$ anchors. By computing the intersection of paths originating from these roots, we isolate valid attribute candidates and apply semantic scoring to select the metric that best aligns with the user's intent.\textbf{(2) Attribute-Driven Reverse Traversal.} Conversely, when the query specifies a target metric but lacks precise context (e.g., \textit{"When did R\&D exceed 100 million?"}), the system executes a bottom-up search. The identified Attribute $a$ acts as a pivot. We back-trace its incoming edges to retrieve all parent Subject and Temporal anchors, effectively enumerating the valid contexts for that specific attribute.Crucially, these two pathways operate in synchrony. The system cross-references the top-down candidates with the bottom-up lineages to validate structural consistency. This intersection mechanism effectively filters out incomplete paths ("orphan nodes"), allowing the system to converge on the optimal $(s, t, a)$ subgraph that maximizes semantic alignment across all dimensions.

In complex reasoning, retrieving semantical information only may not be sufficient. Unlike existing multi-hop reasoning methods that search in a large span space with high requirement in LLM reasoning, we leverage the pre-defined tree structure to efficiently narrow down the search scope.  More specifically, we implement a \textbf{Structural Neighbor Expansion} mechanism. Upon identifying a valid fact tuple $\langle s_i, t_j, a_k \rangle$, the system inspects its structural siblings within the graph topology. A LLM-based analysis model governs this expansion: for example, if the query implies a temporal comparison (e.g., "growth rate"), the system expands along the Temporal Sub-tree to retrieve adjacent time nodes ($t_{j-1}, t_{j+1}$). This ensures the retrieved subgraph captures not just the focal answer, and LLM is guided only to find answer out of a few options, hence the accuracy and efficiency can be maintained.

\subsubsection{Phase II: Text-Bridged Augmentation}

Following the graph navigation phase, the system obtains a set of structured answers $\mathcal{A} = \{ A_i \mid A_i = (s_i, t_i, a_i, v_i, o_i) \}$, where $o_i$ denotes the identifier of the original source table. While these tuples provide high-precision numerical evidence, tabular data in real-world documents is rarely standalone. In professional reports, tables are typically accompanied by narrative texts that offer qualitative insights, causal explanations, or condition caveats. Information in contextual text is complementary to the quantitative cells. To reconstruct this holistic view, we bridge the gap between structured facts and unstructured narratives through two steps as described below.

\textbf{(1) Template-Based Linearization.}
First, we explicitly serialize each retrieved graph tuple $A_i$ into a natural language statement using a deterministic template $\mathbf{T}_i$:
$$ \texttt{"[Subject]'s [Attribute] is [Value] at [Temporal]"} $$
This transformation does not merely format data; it serves as a bridge to translate rigid table entries into a fluent linguistic query compatible with textual search spaces for completing step {(2)}.

\textbf{(2) Cross-Modal Context Fusion.}
Subsequently, we utilize the linearized sentence $\mathbf{T}_i$ as a query to perform embedding-based retrieval on the raw textual corpus associated with the source table $o_i$. By concatenating these top-ranked textual contexts with the structured fact, we achieve a multi-path retrieval strategy that ensures the LLM receives a complete information package: the precise metric from the table and the surrounding analysis from the text.
\begin{table}[t]
\centering
\small
\renewcommand{\arraystretch}{0.9}
\setlength{\tabcolsep}{6pt}
\caption{Pure Tabular Retrieval Performance (\%)} 
\label{tab:pure_retrieval}
\resizebox{\columnwidth}{!}{
\begin{tabular}{l|c|cccc}
\toprule
\textbf{Method} & \textbf{Metric} & \textbf{top-1} & \textbf{top-3} & \textbf{top-5} & \textbf{top-10} \\
\midrule
Standard RAG & HR & 36.6 & 51.9 & 58.9 & 68.9 \\
             & R  & 25.5 & 36.5 & 41.2 & 48.7 \\
             & P  & 36.6 & 17.5 & 12.0 & 7.2  \\
\midrule
LangChain    & HR & 38.2 & 53.5 & 60.3 & 69.3 \\
             & R  & 26.6 & 37.5 & 42.2 & 48.9 \\
             & P  & 38.2 & 18.1 & 12.3 & 7.2  \\
\midrule
Self-RAG     & HR & 11.9 & 19.6 & 25.5 & 38.2 \\
(13B)        & R  & 8.2  & 13.5 & 17.7 & 27.2 \\
             & P  & 11.9 & 6.8  & 5.4  & 4.1  \\
\midrule
MixRAG       & HR & 71.6 & 74.2 & 75.0 & \underline{75.7} \\
             & R  & 59.5 & 65.3 & 68.4 & \textbf{72.4} \\
             & P  & 71.6 & 67.9 & 63.0 & \underline{50.5} \\
\cmidrule(lr){2-6}
             & C-HR & 30.1 & 41.7 & 42.9 & 55.6 \\
             & C-R  & 9.8  & 20.3 & 25.4 & 32.2 \\
             & C-P  & 25.6 & 17.8 & 13.1 & 8.2  \\
\midrule
\midrule
\textbf{FT-RAG} & HR & \multicolumn{4}{c}{\textbf{93.7}} \\
\textbf{(Ours)} & R  & \multicolumn{4}{c}{\underline{70.1}} \\
                & P  & \multicolumn{4}{c}{\textbf{82.6}} \\
\cmidrule(lr){2-6}
                & \textbf{C-HR} & \multicolumn{4}{c}{\textbf{88.5}} \\
                & \textbf{C-R}  & \multicolumn{4}{c}{\textbf{62.0}} \\
                & \textbf{C-P}  & \multicolumn{4}{c}{\textbf{81.9}} \\
\bottomrule
\end{tabular}%
}
\end{table}

\section{Experiments}
\subsection{Experiment Settings}
\subsubsection{Datasets and Task Protocols}
We evaluate the performance of FT-RAG on our curated \textsc{Multi-Table-RAG-Lib} benchmark. To rigorously distinguish between RAG capabilities on structured data versus heterogeneous documents, we design two distinct QA tasks, each containing 4,935 samples:
\begin{itemize}
\item \textbf{Pure Tabular Retrieval}: This setting evaluates the system's ability to extract information exclusively from structured tables. Queries in this subset rely solely on tabular facts without auxiliary textual context.
\item \textbf{Context-Augmented Retrieval}: This setting simulates real-world document QA, where tabular facts must be synthesized with surrounding textual narratives. Each query is intrinsically anchored to surrounding textual segments, requiring the model to capture a complete evidence chain.
\end{itemize}
\subsubsection{Baseline Methods}To comprehensively evaluate effectiveness, we benchmark FT-RAG against several representative frameworks. To ensure a fair comparison, tabular data across all baselines is standardized into HTML5 format to preserve structural semantics \cite{Tan_2025}.\begin{itemize}[nosep]\item \textbf{Standard RAG}: A canonical baseline that treats linearized tables as standard text segments. It uses pre-trained embedding models to construct a dense vector space and retrieves top-$k$ chunks without table-specific optimization.\item \textbf{LangChain} \cite{202411.0566}: A widely adopted pipeline that summarizes row-column content using a multi-vector retriever. It divides long tables into child chunks while preserving links to parent documents to support fine-grained retrieval.\item \textbf{Self-RAG} \cite{asai2023selfraglearningretrievegenerate}: An adaptive framework employing a self-trained critic model. It generates reflection tokens to dynamically evaluate the relevance of retrieved chunks, iteratively optimizing output via a self-feedback loop.\item \textbf{MixRAG} \cite{zhang-2025}: A three-stage framework optimized for heterogeneous RAG. It uses an H-RCL strategy for table structure representation, an LLM-based reranker for alignment, and a multi-step reasoning prompt. Unlike the other methods, MixRAG decomposes tables and converts cells into natural language summaries.\end{itemize}Implementation details are provided in Appendix \ref{appendix:implementation}.

\subsubsection{Evaluation Metrics} \label{sec:metrics}
We employ a comprehensive set of metrics to assess both retrieval and generation quality.\paragraph{Retrieval Metrics}We report \textbf{Hit Rate (HR)}, \textbf{Recall (R)}, and \textbf{Precision (P)} to measure the accuracy of evidence retrieval. Given the distinct nature of the two tasks, we adopt task-specific cutoff thresholds:\begin{itemize}\item For \textbf{Pure Tabular Retrieval}, we report performance at standard cutoffs $k \in \{1, 3, 5, 10\}$.\item For \textbf{Context-Augmented Retrieval}, where the evidence space is larger, we employ a scaled metric set $k \in \{4, 12, 20, 40\}$ to rigorously evaluate the system's ability to capture the complete evidence set.\end{itemize}To further scrutinize the model's capability in fine-grained table understanding, we introduce Cell-level Metrics: Cell-Hit Rate (C-HR), Cell-Recall (C-R), and Cell-Precision (C-P). These metrics quantify the alignment at table cell granularity, independent of the evaluation metrics used for textual narratives.

\paragraph{Generation Metrics} To ensure a balanced evaluation of three metrics, we adopt the Top-5 and Top-20 pure table retrieval results as representative performance for baselines in the pure tabular and context-augmented settings, respectively.  For generation quality, we utilize the Claimify framework \cite{metropolitansky2025effectiveextractionevaluationfactual} to decompose responses into atomic statements. We measure the recall of Exact Value Matching and the precision\&recall of Claim-Level Alignment, ensuring faithful reasoning over tabular data. Details are provided in Appendix \ref{appendix:metrics}.
\begin{table}[t]
\centering
\small
\renewcommand{\arraystretch}{0.9}
\setlength{\tabcolsep}{6pt}
\caption{Context-Augmented Retrieval Performance (\%)} 
\label{tab:context_retrieval}
\resizebox{\columnwidth}{!}{%
\begin{tabular}{l|c|cccc}
\toprule
\textbf{Method} & \textbf{Metric} & \textbf{top-4} & \textbf{top-12} & \textbf{top-20} & \textbf{top-40} \\
\midrule
Standard RAG & HR & 48.5 & 69.6 & 78.6 & 89.0 \\
             & R  & 35.3 & 50.9 & 58.2 & 68.3 \\
             & P  & 30.7 & 18.9 & 12.7 & 6.6  \\
\midrule
LangChain    & HR & 70.74 & 84.24 & 88.57 & \underline{93.27} \\
             & R  & 25.65 & 38.50 & 42.99 & 49.67 \\
             & P  & 70.74 & 43.22 & 31.00 & 19.60 \\
\midrule
Self-RAG     & HR & 67.8 & 82.9 & 87.6 & 89.3 \\
(13B)        & R  & 18.1 & 33.8 & 40.5 & 44.4 \\
             & P  & 67.8 & 44.1 & 31.8 & \underline{24.9} \\
\midrule
MixRAG       & HR & 75.3 & 86.6 & 90.6 & \textbf{95.4} \\
             & R  & 63.6 & 74.7 & 79.3 & \textbf{86.3} \\
             & P  & 30.7 & 14.4 & 9.0  & 4.7  \\
\cmidrule(lr){2-6}
             & C-HR & 56.5 & 58.5 & 58.8 & 59.0 \\
             & C-R  & 11.5 & 17.1 & 20.4 & 25.9 \\
             & C-P  & 24.1 & 12.1 & 8.7  & 5.7  \\
\midrule
\midrule
\textbf{FT-RAG} & HR & \multicolumn{4}{c}{88.3} \\
\textbf{(Ours)} & R  & \multicolumn{4}{c}{\underline{75.0}} \\
                & P  & \multicolumn{4}{c}{\textbf{78.9}} \\
\cmidrule(lr){2-6}
                & C-HR & \multicolumn{4}{c}{\textbf{88.3}} \\
                & C-R  & \multicolumn{4}{c}{\textbf{62.7}} \\
                & C-P  & \multicolumn{4}{c}{\textbf{76.1}} \\
\bottomrule
\end{tabular}
}
\end{table}
\subsection{Main Results}

\subsubsection{Retrieval Performance Analysis} We present the retrieval performance for Pure Tabular and Context-Augmented tasks in Table~\ref{tab:pure_retrieval} and Table~\ref{tab:context_retrieval}, respectively.

In the Pure Tabular setting, results demonstrate that FT-RAG achieves superior performance over all baselines, characterized by a significant increase in retrieval precision. Unlike baseline methods, which suffer from severe precision degradation to maintain Hit Rate and recall, which wasting computational resources on high-volume samplinh, our framework leverages the structural constraints of the SAT-Graph to target evidence precisely. Furthermore, our decisive advantage in Cell-level metrics confirms that the proposed topology effectively preserves the integrity of the evidence chain, enabling LLMs to comprehend tables at a fine-grained level.

In the Context-Augmented setting, the advantage of the cell-level approach becomes even more pronounced. While baselines observe a marginal performance boost due to their proficiency in processing textual chunks, they struggle to maintain structural focus. For instance, MixRAG suffers a notable decrease in Cell-Recall and Cell-Precision, indicating that the added text acts as noise that distracts the model from fine-grained search. In contrast, FT-RAG maintains a consistent superiority by establishing a table-first retrieval hierarchy. By utilizing high-confidence tabular cells as anchors to locate auxiliary textual context, our pipeline effectively shifts the paradigm from a noisy query-seeking-answer search to a stable answer-seeking-evidence process. This mechanism ensures that textual retrieval is strictly grounded in structural facts, thereby mitigating interference and maximizing precision. 

\subsubsection{Generation Alignment and Value Precision} The generation results in Table~\ref{tab:generation_results} reveal a critical distinction between thematic relevance and data fidelity. While Standard RAG achieves the highest Claim Precision, its substantially lower Value Accuracy Recall exposes a severe semantic misalignment. Conventional baseline approaches often retrieve the correct topic but fail to ground specific numerical attributes accurately. In contrast, FT-RAG demonstrates outperforming numerical comprehension, achieving a dominant Value Accuracy of 54.5\%, alongside the highest Claim Recall. This confirms that by anchoring retrieval to the SAT-Graph, our method effectively locks values to their precise semantic coordinates.
\begin{table}[h]
\centering
\caption{Generation Performance Comparison}
\label{tab:generation_results}
\resizebox{\linewidth}{!}{%
\begin{tabular}{l|c|cc} 
\toprule
\textbf{Method} & \textbf{Value Accuracy Recall}& \textbf{Claim Precision} & \textbf{Claim Recall} \\
\midrule
Standard RAG   & 22.0\% & \textbf{60.5\%} & \underline{55.8\%} \\
LangChain      & 24.2\% & 51.3\% & 51.0\% \\
Self-RAG       & 18.1\% & 48.2\% & 24.9\% \\
MixRAG         & \underline{33.9\%} & 41.8\% & 48.9\% \\
\midrule
\textbf{FT-RAG (Ours)} & \textbf{54.5\%} & \underline{59.0\%} & \textbf{70.6\%} \\
\bottomrule
\end{tabular}%
}
\end{table}
\subsection{Ablation Study}
To rigorously evaluate the architecture of FT-RAG, we conducted an ablation study to quantify the individual contributions of its core components. Given that our methodology represents a systemic departure from conventional RAG paradigms, we assessed the performance impact by selectively disabling the following modules:
\begin{itemize}

\item\textbf{w/o SAT-Graph}: This variant bypasses the graph-based navigation layer, reverting to a standard two-step chunk-based retrieval to measure the structural advantage of the SAT-Graph.

\item\textbf{w/o Structural Neighbor Expansion mechanism}: This configuration disables the mechanism that associates neighboring nodes during retrieval, allowing us to isolate the importance of cell-level contextual connectivity.

\item\textbf{w/o Multi-modal Fusion Module}: In this setup, we omit the integration of surrounding textual context, evaluating the module's effectiveness in augmenting tabular evidence with narrative data.

\item\textbf{FT-RAG}: The complete set of our approach.
\end{itemize}

To ensure consistency and comparative rigor, all variants were evaluated based on their context-augmented generation performance, serving as a proxy for the significance of each architectural component in the overall retrieval-reasoning pipeline.

\begin{figure}[htbp]
    \centering
    \begin{tikzpicture}
    \begin{axis}[
        width=\linewidth, 
        height=6cm,     
        grid=major,       
        grid style={dashed, gray!30}, 
        ylabel={Score (\%)},
        ylabel style={font=\small},
        xlabel style={font=\small},
        symbolic x coords={w/o SAT, w/o SNE, w/o Fusion, FT-RAG},
        xtick=data,
        xticklabel style={font=\scriptsize, rotate=20, anchor=north east},
        ymin=20, ymax=60, 
        legend style={
            at={(0.02,0.98)}, 
            anchor=north west, 
            font=\scriptsize, 
            cells={anchor=west}, 
            fill=white, 
            fill opacity=0.8,
            draw opacity=0.5
        },
        every axis plot/.append style={very thick, mark size=2pt},
        nodes near coords,
        every node near coord/.append style={
            font=\tiny, 
            anchor=south, 
            /pgf/number format/fixed,
            /pgf/number format/precision=1
        },
    ]
    
    \addplot[color1, mark=*] coordinates {
        (w/o SAT, 26.3) (w/o SNE, 48.2) (w/o Fusion, 40.3) (FT-RAG, 51.6)
    };
    \addlegendentry{Value Acc.}
    
    \addplot[color2, mark=square*] coordinates {
        (w/o SAT, 31.8) (w/o SNE, 42.1) (w/o Fusion, 39.0) (FT-RAG, 41.2)
    };
    \addlegendentry{Claim Prec.}
    
    \addplot[color3, mark=triangle*] coordinates {
        (w/o SAT, 36.7) (w/o SNE, 47.2) (w/o Fusion, 44.7) (FT-RAG, 52.3)
    };
    \addlegendentry{Claim Rec.}
    
    \end{axis}
    \end{tikzpicture}
    \caption{Performance across ablation settings. Here, w/o SAT, w/o SNE, and w/o Fusion denote the removal of SAT-Graph, Structural Neighbor Expansion, and Multi-modal Fusion Module, respectively. }
    \label{fig:ablation_line}
\end{figure}
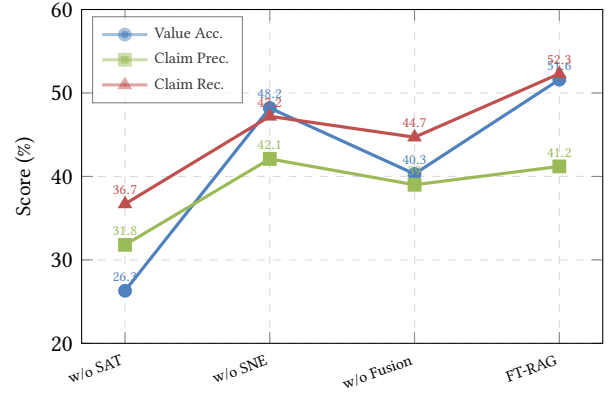
As illustrated in Figure \ref{fig:ablation_line}, the ablation results validate the efficacy of FT-RAG's architectural components. Our analysis yields three key observations: i) \textbf{The SAT-Graph} construction is foundational; removing this module causes all metric scores to plummet below 40\%. This underscores the limitation of traditional chunk-level retrieval, which lacks the fine-grained decomposition necessary to capture precise information from tabular and textual data. ii) \textbf{The Structural Neighbor Expansion mechanism} have relatively marginal gains compared to other components, but it enhances the comprehensiveness of evidence grounding. While expanding the search scope incurs a marginal decline in Claim Precision due to increased context, the overall performance gain in FT-RAG confirms its critical role in bridging informational gaps for complex queries. iii) \textbf{The Multi-modal Fusion module} is vital not only for semantic completeness but also for precise value extraction. Notably, the exclusion of this module leads to a dramatic decrease in Value Accuracy. This finding corroborates the hypothesis that integrating surrounding textual context in table-centric QA tasks helps LLMs to accurately interpret and disambiguate tabular values.

\section{Conclusion}

In this paper, we address the challenge of complex reasoning in Table-Centric QA, specifically targeting scenarios involving intricate tabular data. To this end, we propose FT-RAG, a novel framework that redefines tabular retrieval by operating at the fine-grained cell level. By decomposing tables into semantic units and organizing them into a unified SAT graph, FT-RAG enables direct semantic alignment between queries and individual cells. This structure supports a multi-modal fusion strategy that connect retrieved evidence with surrounding textual context to create grounded evidence. To facilitate this research, we also introduce Multi-Table-RAG-Lib, a comprehensive dataset for multi-table relational querying. Extensive experiments demonstrate that FT-RAG consistently outperforms existing baselines, establishing a new state-of-the-art for table-centric RAG tasks. A detailed discussion on limitations and the future work of our approach is provided in Appendix \ref{appendix:dal}. 
\section*{Acknowledgement}

The author acknowledges the use of Google Antigravity in the development and optimization of data processing scripts.
\clearpage
\bibliographystyle{ACM-Reference-Format} 
\bibliography{references.bib}


\begin{thebibliography}{28}


\ifx \showCODEN    \undefined \def \showCODEN     #1{\unskip}     \fi
\ifx \showISBNx    \undefined \def \showISBNx     #1{\unskip}     \fi
\ifx \showISBNxiii \undefined \def \showISBNxiii  #1{\unskip}     \fi
\ifx \showISSN     \undefined \def \showISSN      #1{\unskip}     \fi
\ifx \showLCCN     \undefined \def \showLCCN      #1{\unskip}     \fi
\ifx \shownote     \undefined \def \shownote      #1{#1}          \fi
\ifx \showarticletitle \undefined \def \showarticletitle #1{#1}   \fi
\ifx \showURL      \undefined \def \showURL       {\relax}        \fi
\providecommand\bibfield[2]{#2}
\providecommand\bibinfo[2]{#2}
\providecommand\natexlab[1]{#1}
\providecommand\showeprint[2][]{arXiv:#2}

\bibitem[Arslan et~al\mbox{.}(2024)]%
        {arslan-2024}
\bibfield{author}{\bibinfo{person}{Muhammad Arslan}, \bibinfo{person}{Hussam Ghanem}, \bibinfo{person}{Saba Munawar}, {and} \bibinfo{person}{Christophe Cruz}.} \bibinfo{year}{2024}\natexlab{}.
\newblock \showarticletitle{{A Survey on RAG with LLMs}}.
\newblock \bibinfo{journal}{\emph{Procedia Computer Science}}  \bibinfo{volume}{246} (\bibinfo{date}{1} \bibinfo{year}{2024}), \bibinfo{pages}{3781--3790}.
\newblock
\href{https://doi.org/10.1016/j.procs.2024.09.178}{doi:\nolinkurl{10.1016/j.procs.2024.09.178}}


\bibitem[Asai et~al\mbox{.}(2023)]%
        {asai2023selfraglearningretrievegenerate}
\bibfield{author}{\bibinfo{person}{Akari Asai}, \bibinfo{person}{Zeqiu Wu}, \bibinfo{person}{Yizhong Wang}, \bibinfo{person}{Avirup Sil}, {and} \bibinfo{person}{Hannaneh Hajishirzi}.} \bibinfo{year}{2023}\natexlab{}.
\newblock \bibinfo{title}{Self-RAG: Learning to Retrieve, Generate, and Critique through Self-Reflection}.
\newblock
\showeprint[arxiv]{2310.11511}~[cs.CL]
\urldef\tempurl%
\url{https://arxiv.org/abs/2310.11511}
\showURL{%
\tempurl}


\bibitem[Fan et~al\mbox{.}(2024)]%
        {fan-2024}
\bibfield{author}{\bibinfo{person}{Wenqi Fan}, \bibinfo{person}{Yujuan Ding}, \bibinfo{person}{Liangbo Ning}, \bibinfo{person}{Shijie Wang}, \bibinfo{person}{Hengyun Li}, \bibinfo{person}{Dawei Yin}, \bibinfo{person}{Tat-Seng Chua}, {and} \bibinfo{person}{Qing Li}.} \bibinfo{year}{2024}\natexlab{}.
\newblock \showarticletitle{{A Survey on RAG Meeting LLMs: Towards Retrieval-Augmented Large Language Models}}.
\newblock \bibinfo{journal}{\emph{KDD '24: Proceedings of the 30th ACM SIGKDD Conference on Knowledge Discovery and Data Mining}} (\bibinfo{date}{8} \bibinfo{year}{2024}), \bibinfo{pages}{6491--6501}.
\newblock
\href{https://doi.org/10.1145/3637528.3671470}{doi:\nolinkurl{10.1145/3637528.3671470}}


\bibitem[Guo et~al\mbox{.}(2025a)]%
        {guo-2025}
\bibfield{author}{\bibinfo{person}{Zirui Guo}, \bibinfo{person}{Xubin Ren}, \bibinfo{person}{Lingrui Xu}, \bibinfo{person}{Jiahao Zhang}, {and} \bibinfo{person}{Chao Huang}.} \bibinfo{year}{2025}\natexlab{a}.
\newblock \bibinfo{title}{{RAG-Anything: All-in-One RAG framework}}.
\newblock
\urldef\tempurl%
\url{https://arxiv.org/abs/2510.12323}
\showURL{%
\tempurl}


\bibitem[Guo et~al\mbox{.}(2025b)]%
        {guo2025lightragsimplefastretrievalaugmented}
\bibfield{author}{\bibinfo{person}{Zirui Guo}, \bibinfo{person}{Lianghao Xia}, \bibinfo{person}{Yanhua Yu}, \bibinfo{person}{Tu Ao}, {and} \bibinfo{person}{Chao Huang}.} \bibinfo{year}{2025}\natexlab{b}.
\newblock \bibinfo{title}{LightRAG: Simple and Fast Retrieval-Augmented Generation}.
\newblock
\showeprint[arxiv]{2410.05779}~[cs.IR]
\urldef\tempurl%
\url{https://arxiv.org/abs/2410.05779}
\showURL{%
\tempurl}


\bibitem[Gutiérrez et~al\mbox{.}(2025)]%
        {gutierrez2025hipporagneurobiologicallyinspiredlongterm}
\bibfield{author}{\bibinfo{person}{Bernal~Jiménez Gutiérrez}, \bibinfo{person}{Yiheng Shu}, \bibinfo{person}{Yu Gu}, \bibinfo{person}{Michihiro Yasunaga}, {and} \bibinfo{person}{Yu Su}.} \bibinfo{year}{2025}\natexlab{}.
\newblock \bibinfo{title}{HippoRAG: Neurobiologically Inspired Long-Term Memory for Large Language Models}.
\newblock
\showeprint[arxiv]{2405.14831}~[cs.CL]
\urldef\tempurl%
\url{https://arxiv.org/abs/2405.14831}
\showURL{%
\tempurl}


\bibitem[Kang et~al\mbox{.}(2025)]%
        {kang-etal-2025-grit}
\bibfield{author}{\bibinfo{person}{Yujin Kang}, \bibinfo{person}{Park~Seong Woo}, {and} \bibinfo{person}{Yoon-Sik Cho}.} \bibinfo{year}{2025}\natexlab{}.
\newblock \showarticletitle{{GRIT}: Guided Relational Integration for Efficient Multi-Table Understanding}. In \bibinfo{booktitle}{\emph{Proceedings of the 2025 Conference on Empirical Methods in Natural Language Processing}}, \bibfield{editor}{\bibinfo{person}{Christos Christodoulopoulos}, \bibinfo{person}{Tanmoy Chakraborty}, \bibinfo{person}{Carolyn Rose}, {and} \bibinfo{person}{Violet Peng}} (Eds.). \bibinfo{publisher}{Association for Computational Linguistics}, \bibinfo{address}{Suzhou, China}, \bibinfo{pages}{21984--21997}.
\newblock
\showISBNx{979-8-89176-332-6}
\href{https://doi.org/10.18653/v1/2025.emnlp-main.1118}{doi:\nolinkurl{10.18653/v1/2025.emnlp-main.1118}}


\bibitem[Lewis et~al\mbox{.}(2021)]%
        {lewis2021retrievalaugmentedgenerationknowledgeintensivenlp}
\bibfield{author}{\bibinfo{person}{Patrick Lewis}, \bibinfo{person}{Ethan Perez}, \bibinfo{person}{Aleksandra Piktus}, \bibinfo{person}{Fabio Petroni}, \bibinfo{person}{Vladimir Karpukhin}, \bibinfo{person}{Naman Goyal}, \bibinfo{person}{Heinrich Küttler}, \bibinfo{person}{Mike Lewis}, \bibinfo{person}{Wen tau Yih}, \bibinfo{person}{Tim Rocktäschel}, \bibinfo{person}{Sebastian Riedel}, {and} \bibinfo{person}{Douwe Kiela}.} \bibinfo{year}{2021}\natexlab{}.
\newblock \bibinfo{title}{Retrieval-Augmented Generation for Knowledge-Intensive NLP Tasks}.
\newblock
\showeprint[arxiv]{2005.11401}~[cs.CL]
\urldef\tempurl%
\url{https://arxiv.org/abs/2005.11401}
\showURL{%
\tempurl}


\bibitem[Lin(2004)]%
        {lin-2004-rouge}
\bibfield{author}{\bibinfo{person}{Chin-Yew Lin}.} \bibinfo{year}{2004}\natexlab{}.
\newblock \showarticletitle{{ROUGE}: A Package for Automatic Evaluation of Summaries}. In \bibinfo{booktitle}{\emph{Text Summarization Branches Out}}. \bibinfo{publisher}{Association for Computational Linguistics}, \bibinfo{address}{Barcelona, Spain}, \bibinfo{pages}{74--81}.
\newblock
\urldef\tempurl%
\url{https://aclanthology.org/W04-1013/}
\showURL{%
\tempurl}


\bibitem[Liu et~al\mbox{.}(2026)]%
        {liu2026graphsearchagenticsearchaugmentedreasoning}
\bibfield{author}{\bibinfo{person}{Jiajin Liu}, \bibinfo{person}{Yuanfu Sun}, \bibinfo{person}{Dongzhe Fan}, {and} \bibinfo{person}{Qiaoyu Tan}.} \bibinfo{year}{2026}\natexlab{}.
\newblock \bibinfo{title}{GraphSearch: Agentic Search-Augmented Reasoning for Zero-Shot Graph Learning}.
\newblock
\showeprint[arxiv]{2601.08621}~[cs.CL]
\urldef\tempurl%
\url{https://arxiv.org/abs/2601.08621}
\showURL{%
\tempurl}


\bibitem[Luo et~al\mbox{.}(2023)]%
        {luo2023hrothybridpromptstrategy}
\bibfield{author}{\bibinfo{person}{Tongxu Luo}, \bibinfo{person}{Fangyu Lei}, \bibinfo{person}{Jiahe Lei}, \bibinfo{person}{Weihao Liu}, \bibinfo{person}{Shihu He}, \bibinfo{person}{Jun Zhao}, {and} \bibinfo{person}{Kang Liu}.} \bibinfo{year}{2023}\natexlab{}.
\newblock \bibinfo{title}{HRoT: Hybrid prompt strategy and Retrieval of Thought for Table-Text Hybrid Question Answering}.
\newblock
\showeprint[arxiv]{2309.12669}~[cs.CL]
\urldef\tempurl%
\url{https://arxiv.org/abs/2309.12669}
\showURL{%
\tempurl}


\bibitem[Mavroudis(2024)]%
        {202411.0566}
\bibfield{author}{\bibinfo{person}{Vasilios Mavroudis}.} \bibinfo{year}{2024}\natexlab{}.
\newblock \showarticletitle{LangChain v0.3}.
\newblock \bibinfo{journal}{\emph{Preprints}} (\bibinfo{date}{November} \bibinfo{year}{2024}).
\newblock
\href{https://doi.org/10.20944/preprints202411.0566.v1}{doi:\nolinkurl{10.20944/preprints202411.0566.v1}}


\bibitem[Metropolitansky and Larson(2025)]%
        {metropolitansky2025effectiveextractionevaluationfactual}
\bibfield{author}{\bibinfo{person}{Dasha Metropolitansky} {and} \bibinfo{person}{Jonathan Larson}.} \bibinfo{year}{2025}\natexlab{}.
\newblock \bibinfo{title}{Towards Effective Extraction and Evaluation of Factual Claims}.
\newblock
\showeprint[arxiv]{2502.10855}~[cs.CL]
\urldef\tempurl%
\url{https://arxiv.org/abs/2502.10855}
\showURL{%
\tempurl}


\bibitem[Robertson and Zaragoza(2009)]%
        {10.1561/1500000019}
\bibfield{author}{\bibinfo{person}{Stephen Robertson} {and} \bibinfo{person}{Hugo Zaragoza}.} \bibinfo{year}{2009}\natexlab{}.
\newblock \showarticletitle{The Probabilistic Relevance Framework: BM25 and Beyond}.
\newblock \bibinfo{journal}{\emph{Found. Trends Inf. Retr.}} \bibinfo{volume}{3}, \bibinfo{number}{4} (\bibinfo{date}{April} \bibinfo{year}{2009}), \bibinfo{pages}{333–389}.
\newblock
\showISSN{1554-0669}
\href{https://doi.org/10.1561/1500000019}{doi:\nolinkurl{10.1561/1500000019}}


\bibitem[Sarthi et~al\mbox{.}(2024)]%
        {sarthi2024raptorrecursiveabstractiveprocessing}
\bibfield{author}{\bibinfo{person}{Parth Sarthi}, \bibinfo{person}{Salman Abdullah}, \bibinfo{person}{Aditi Tuli}, \bibinfo{person}{Shubh Khanna}, \bibinfo{person}{Anna Goldie}, {and} \bibinfo{person}{Christopher~D. Manning}.} \bibinfo{year}{2024}\natexlab{}.
\newblock \bibinfo{title}{RAPTOR: Recursive Abstractive Processing for Tree-Organized Retrieval}.
\newblock
\showeprint[arxiv]{2401.18059}~[cs.CL]
\urldef\tempurl%
\url{https://arxiv.org/abs/2401.18059}
\showURL{%
\tempurl}


\bibitem[Tan et~al\mbox{.}(2025)]%
        {Tan_2025}
\bibfield{author}{\bibinfo{person}{Jiejun Tan}, \bibinfo{person}{Zhicheng Dou}, \bibinfo{person}{Wen Wang}, \bibinfo{person}{Mang Wang}, \bibinfo{person}{Weipeng Chen}, {and} \bibinfo{person}{Ji-Rong Wen}.} \bibinfo{year}{2025}\natexlab{}.
\newblock \showarticletitle{HtmlRAG: HTML is Better Than Plain Text for Modeling Retrieved Knowledge in RAG Systems}. In \bibinfo{booktitle}{\emph{Proceedings of the ACM on Web Conference 2025}} \emph{(\bibinfo{series}{WWW ’25})}. \bibinfo{publisher}{ACM}, \bibinfo{pages}{1733–1746}.
\newblock
\href{https://doi.org/10.1145/3696410.3714546}{doi:\nolinkurl{10.1145/3696410.3714546}}


\bibitem[Tang and Yang(2024)]%
        {tang2024multihopragbenchmarkingretrievalaugmentedgeneration}
\bibfield{author}{\bibinfo{person}{Yixuan Tang} {and} \bibinfo{person}{Yi Yang}.} \bibinfo{year}{2024}\natexlab{}.
\newblock \bibinfo{title}{MultiHop-RAG: Benchmarking Retrieval-Augmented Generation for Multi-Hop Queries}.
\newblock
\showeprint[arxiv]{2401.15391}~[cs.CL]
\urldef\tempurl%
\url{https://arxiv.org/abs/2401.15391}
\showURL{%
\tempurl}


\bibitem[Wang et~al\mbox{.}(2024)]%
        {wang2024mineruopensourcesolutionprecise}
\bibfield{author}{\bibinfo{person}{Bin Wang}, \bibinfo{person}{Chao Xu}, \bibinfo{person}{Xiaomeng Zhao}, \bibinfo{person}{Linke Ouyang}, \bibinfo{person}{Fan Wu}, \bibinfo{person}{Zhiyuan Zhao}, \bibinfo{person}{Rui Xu}, \bibinfo{person}{Kaiwen Liu}, \bibinfo{person}{Yuan Qu}, \bibinfo{person}{Fukai Shang}, \bibinfo{person}{Bo Zhang}, \bibinfo{person}{Liqun Wei}, \bibinfo{person}{Zhihao Sui}, \bibinfo{person}{Wei Li}, \bibinfo{person}{Botian Shi}, \bibinfo{person}{Yu Qiao}, \bibinfo{person}{Dahua Lin}, {and} \bibinfo{person}{Conghui He}.} \bibinfo{year}{2024}\natexlab{}.
\newblock \bibinfo{title}{MinerU: An Open-Source Solution for Precise Document Content Extraction}.
\newblock
\showeprint[arxiv]{2409.18839}~[cs.CV]
\urldef\tempurl%
\url{https://arxiv.org/abs/2409.18839}
\showURL{%
\tempurl}


\bibitem[Wu et~al\mbox{.}(2025)]%
        {wu2025tablebenchcomprehensivecomplexbenchmark}
\bibfield{author}{\bibinfo{person}{Xianjie Wu}, \bibinfo{person}{Jian Yang}, \bibinfo{person}{Linzheng Chai}, \bibinfo{person}{Ge Zhang}, \bibinfo{person}{Jiaheng Liu}, \bibinfo{person}{Xinrun Du}, \bibinfo{person}{Di Liang}, \bibinfo{person}{Daixin Shu}, \bibinfo{person}{Xianfu Cheng}, \bibinfo{person}{Tianzhen Sun}, \bibinfo{person}{Guanglin Niu}, \bibinfo{person}{Tongliang Li}, {and} \bibinfo{person}{Zhoujun Li}.} \bibinfo{year}{2025}\natexlab{}.
\newblock \bibinfo{title}{TableBench: A Comprehensive and Complex Benchmark for Table Question Answering}.
\newblock
\showeprint[arxiv]{2408.09174}~[cs.CL]
\urldef\tempurl%
\url{https://arxiv.org/abs/2408.09174}
\showURL{%
\tempurl}


\bibitem[Yu et~al\mbox{.}(2025)]%
        {yu-2025}
\bibfield{author}{\bibinfo{person}{Xiaohan Yu}, \bibinfo{person}{Pu Jian}, {and} \bibinfo{person}{Chong Chen}.} \bibinfo{year}{2025}\natexlab{}.
\newblock \bibinfo{title}{{TableRAG: A Retrieval Augmented generation framework for Heterogeneous document Reasoning}}.
\newblock
\urldef\tempurl%
\url{https://arxiv.org/abs/2506.10380}
\showURL{%
\tempurl}


\bibitem[Zhang et~al\mbox{.}(2025a)]%
        {zhang-2025}
\bibfield{author}{\bibinfo{person}{Chi Zhang}, \bibinfo{person}{Qiyang Chen}, {and} \bibinfo{person}{Mengqi Zhang}.} \bibinfo{year}{2025}\natexlab{a}.
\newblock \bibinfo{title}{{Mixture-of-RAG: Integrating Text and Tables with Large Language Models}}.
\newblock
\urldef\tempurl%
\url{https://arxiv.org/abs/2504.09554}
\showURL{%
\tempurl}


\bibitem[Zhang et~al\mbox{.}(2020)]%
        {DBLP:conf/iclr/ZhangKWWA20}
\bibfield{author}{\bibinfo{person}{Tianyi Zhang}, \bibinfo{person}{Varsha Kishore}, \bibinfo{person}{Felix Wu}, \bibinfo{person}{Kilian~Q. Weinberger}, {and} \bibinfo{person}{Yoav Artzi}.} \bibinfo{year}{2020}\natexlab{}.
\newblock \showarticletitle{BERTScore: Evaluating Text Generation with {BERT}}. In \bibinfo{booktitle}{\emph{8th International Conference on Learning Representations, {ICLR} 2020, Addis Ababa, Ethiopia, April 26-30, 2020}}. \bibinfo{publisher}{OpenReview.net}.
\newblock
\urldef\tempurl%
\url{https://openreview.net/forum?id=SkeHuCVFDr}
\showURL{%
\tempurl}


\bibitem[Zhang et~al\mbox{.}(2025b)]%
        {zhang2025rotenhancingtablereasoning}
\bibfield{author}{\bibinfo{person}{Xuanliang Zhang}, \bibinfo{person}{Dingzirui Wang}, \bibinfo{person}{Keyan Xu}, \bibinfo{person}{Qingfu Zhu}, {and} \bibinfo{person}{Wanxiang Che}.} \bibinfo{year}{2025}\natexlab{b}.
\newblock \bibinfo{title}{RoT: Enhancing Table Reasoning with Iterative Row-Wise Traversals}.
\newblock
\showeprint[arxiv]{2505.15110}~[cs.CL]
\urldef\tempurl%
\url{https://arxiv.org/abs/2505.15110}
\showURL{%
\tempurl}


\bibitem[Zhao et~al\mbox{.}(2022)]%
        {zhao-etal-2022-multihiertt}
\bibfield{author}{\bibinfo{person}{Yilun Zhao}, \bibinfo{person}{Yunxiang Li}, \bibinfo{person}{Chenying Li}, {and} \bibinfo{person}{Rui Zhang}.} \bibinfo{year}{2022}\natexlab{}.
\newblock \showarticletitle{{M}ulti{H}iertt: Numerical Reasoning over Multi Hierarchical Tabular and Textual Data}. In \bibinfo{booktitle}{\emph{Proceedings of the 60th Annual Meeting of the Association for Computational Linguistics (Volume 1: Long Papers)}}, \bibfield{editor}{\bibinfo{person}{Smaranda Muresan}, \bibinfo{person}{Preslav Nakov}, {and} \bibinfo{person}{Aline Villavicencio}} (Eds.). \bibinfo{publisher}{Association for Computational Linguistics}, \bibinfo{address}{Dublin, Ireland}, \bibinfo{pages}{6588--6600}.
\newblock
\href{https://doi.org/10.18653/v1/2022.acl-long.454}{doi:\nolinkurl{10.18653/v1/2022.acl-long.454}}


\bibitem[Zhu et~al\mbox{.}(2025b)]%
        {zhu2025knowledgegraphguidedretrievalaugmented}
\bibfield{author}{\bibinfo{person}{Xiangrong Zhu}, \bibinfo{person}{Yuexiang Xie}, \bibinfo{person}{Yi Liu}, \bibinfo{person}{Yaliang Li}, {and} \bibinfo{person}{Wei Hu}.} \bibinfo{year}{2025}\natexlab{b}.
\newblock \bibinfo{title}{Knowledge Graph-Guided Retrieval Augmented Generation}.
\newblock
\showeprint[arxiv]{2502.06864}~[cs.CL]
\urldef\tempurl%
\url{https://arxiv.org/abs/2502.06864}
\showURL{%
\tempurl}


\bibitem[Zhu et~al\mbox{.}(2025a)]%
        {zhu2025rightanswerrighttime}
\bibfield{author}{\bibinfo{person}{Zulun Zhu}, \bibinfo{person}{Haoyu Liu}, \bibinfo{person}{Mengke He}, {and} \bibinfo{person}{Siqiang Luo}.} \bibinfo{year}{2025}\natexlab{a}.
\newblock \bibinfo{title}{Right Answer at the Right Time - Temporal Retrieval-Augmented Generation via Graph Summarization}.
\newblock
\showeprint[arxiv]{2510.16715}~[cs.IR]
\urldef\tempurl%
\url{https://arxiv.org/abs/2510.16715}
\showURL{%
\tempurl}


\bibitem[Zhuang et~al\mbox{.}(2025)]%
        {zhuang2025linearraglineargraphretrieval}
\bibfield{author}{\bibinfo{person}{Luyao Zhuang}, \bibinfo{person}{Shengyuan Chen}, \bibinfo{person}{Yilin Xiao}, \bibinfo{person}{Huachi Zhou}, \bibinfo{person}{Yujing Zhang}, \bibinfo{person}{Hao Chen}, \bibinfo{person}{Qinggang Zhang}, {and} \bibinfo{person}{Xiao Huang}.} \bibinfo{year}{2025}\natexlab{}.
\newblock \bibinfo{title}{LinearRAG: Linear Graph Retrieval Augmented Generation on Large-scale Corpora}.
\newblock
\showeprint[arxiv]{2510.10114}~[cs.CL]
\urldef\tempurl%
\url{https://arxiv.org/abs/2510.10114}
\showURL{%
\tempurl}


\bibitem[Zou et~al\mbox{.}(2025)]%
        {zou-2025}
\bibfield{author}{\bibinfo{person}{Jiaru Zou}, \bibinfo{person}{Dongqi Fu}, \bibinfo{person}{Sirui Chen}, \bibinfo{person}{Xinrui He}, \bibinfo{person}{Zihao Li}, \bibinfo{person}{Yada Zhu}, \bibinfo{person}{Jiawei Han}, {and} \bibinfo{person}{Jingrui He}.} \bibinfo{year}{2025}\natexlab{}.
\newblock \bibinfo{title}{{RAG over Tables: Hierarchical Memory Index, Multi-Stage Retrieval, and Benchmarking}}.
\newblock
\urldef\tempurl%
\url{https://arxiv.org/abs/2504.01346}
\showURL{%
\tempurl}


\end{thebibliography}

\appendix
\section{Data Preparation}
\label{app:data_prep}

The data construction pipeline consists of two phases: (1) Entity Enrichment for metadata restoration, and (2) Randomized QA Pair Generation with strict bias control.

\subsection{Phase I: Entity Enrichment via LLM Inference}
Since raw \textsc{MultiHiertt} data lacks central subjects due to anonymization, we require a global metadata field to support cell grouping. To restore this without human intervention, we employ a \textbf{Context-to-Entity Extraction} prompt. By feeding the full document context (including titles and intros) into an LLM with a 1-shot demonstration, we accurately infer the specific entity name to serve as the global anchor.

\subsection{Phase II: Unbiased QA Generation Strategy}
To prevent inductive bias that artificially inflates the performance of our semantic-based FT-RAG, we enforce a \textbf{Stochastic Field Association Strategy} that strictly excludes semantic similarity during generation. The process includes:

\begin{enumerate}
    \item \textbf{Global Shuffling:} Tables across the entire corpus are flattened and shuffled to disrupt pre-existing document order.
    \item \textbf{Heuristic Matching:} Target cells are paired solely via hard-coded heuristics (e.g., exact string matches on dates) or random selection. We explicitly avoid vector similarity or embedding distances.
    \item \textbf{Rejection Sampling:} An LLM serves as a logic validator. It assesses randomly paired cells and discards those unable to form a coherent multi-hop question, ensuring the dataset reflects logical rather than semantic proximity.
\end{enumerate}

The specific prompt templates for both phases are illustrated in Figure \ref{fig:app_prompts}.

\section{Dataset Statistics}
\label{app:dataset_stats}

Table \ref{tab:dataset_stats} summarizes the statistical characteristics of the \textsc{Multi-Table-RAG-Lib} subset derived from \textsc{MultiHiertt}. The dataset comprises 81 complex financial and administrative documents, containing a total of 328 tables and nearly 10,000 individual semantic cell units. 
\begin{table}[h]
    \centering
    \footnotesize 
    \setlength{\tabcolsep}{5pt} 
    \renewcommand{\arraystretch}{0.9} 
    
    \caption{\textbf{Statistics of the Multi-Table-RAG-Lib Subset.}}
    \label{tab:dataset_stats}
    
    \begin{tabular}{lr}
    \toprule
    \textbf{Metric} & \textbf{Count} \\
    \midrule
    Total Documents & 81 \\
    Total Tables & 328 \\
    Total Semantic Cell Units & 9,990 \\
    QA Pairs ($f=0$, Pure Retrieval) & 4,935 \\
    QA Pairs ($f=1$, Augmented) & 4,935 \\
    \textbf{Total QA Pairs} & \textbf{9,870} \\
    \midrule
    \multicolumn{2}{l}{\textit{Average Statistics per Document}} \\ 
    \quad Tables / Doc & 4.05 \scriptsize{(Range: 3--7)} \\
    \quad Cells / Doc & 123.33 \scriptsize{(Range: 32--245)} \\
    \quad Cells / Table & 32.44 \\
    \bottomrule
    \end{tabular}
\end{table}

These statistics highlight two critical challenges that motivate our FT-RA* design:
\begin{itemize}
    \item \textbf{High Information Density:} With an average of 123.33 cells per document and some documents containing up to 245 cells, traditional linearization methods would result in excessively long contexts. This validates our Fine-Grained Table Decomposition strategy, which breaks these massive structures into manageable atomic units.
    \item \textbf{Multi-Table Interdependency:} The dataset features an average of 4.05 tables per document (ranging up to 7). This distribution is specifically curated to evaluate Cross-Table Reasoning. It necessitates a graph-based approach (our SAT Graph) to navigate disjointed tabular islands, ensuring that the retrieval mechanism can hop between multiple tables to synthesize a complete answer.
\end{itemize}

\section{Implementation Details}\label{appendix:implementation}
Our experiment leverages an Intel Core i9-13980HX CPU for processing and a local NVIDIA GeForce RTX 4080 Laptop GPU for deploying small-scale model. For models exceeding local VRAM capacity, computational tasks are offloaded to a server-side NVIDIA H100 GPU to ensure efficient inference.  "Regarding the model configurations, we employ BGE-Large-En as the primary embedding backbone for all baseline methods to encode document chunks into dense vector spaces. To address the excessively fine-grained text segmentation in the original \textbackslash{}textsc\{MultiHiertt\} dataset, we implement a heuristic concatenation strategy to aggregate text fragments, ensuring the resulting chunks align with the standard lengths typically found in production-level RAG applications. Furthermore, Qwen3-32B is utilized as a high-capacity LLM expert to synthesize complex question-answer pairs for our benchmark, while Qwen3-Max serves as the authoritative reasoning engine during the final RAG generation phase. We consistently employ the Qwen series across all modules to mitigate potential biases arising from linguistic styles and reasoning behaviors of different LLM architectures.

\section{Detailed Evaluation Metrics}\label{appendix:metrics}

To rigorously evaluate retrieval performance in multi-ground-truth tasks, we employ the following three metrics:

\textbf{(1) Hit Rate (Hit@K)}: Used as a baseline indicator of relevance, measuring if \textit{at least one} ground-truth evidence piece is present in the top-$K$ results:
$$ \text{Hit}@K = \frac{1}{|Q|} \sum_{q \in Q} \mathbb{I}(|R_q^K \cap G_q| \ge 1) $$
where $G_q$ is the ground-truth set and $R_q^K$ represents the top-$K$ retrieved candidates.

\textbf{(2) Recall (Recall@K)}: Quantifies the \textit{coverage} of necessary evidence. In tabular reasoning (e.g., trend analysis), missing a single cell can render the entire context insufficient:
$$ \text{Recall}@K = \frac{1}{|Q|} \sum_{q \in Q} \frac{|R_q^K \cap G_q|}{|G_q|} $$

\textbf{(3) Precision (Precision@K)}: Measures the signal-to-noise ratio to ensure the LLM is not distracted by irrelevant information:
$$ \text{Precision}@K = \frac{1}{|Q|} \sum_{q \in Q} \frac{|R_q^K \cap G_q|}{K} $$

For generation quality, conventional protocols relying on n-gram overlap (e.g., ROUGE\cite{lin-2004-rouge}) or semantic embedding similarity (e.g., BERTScore\cite{DBLP:conf/iclr/ZhangKWWA20}) frequently yield inflated scores for fluent but numerically inaccurate hallucinations. Conversely, rigid metrics such as Exact Match (EM) are insufficient for complex reasoning tasks. To bridge this gap, we implement a granular, claim-level faithfulness evaluation pipeline anchored by \textbf{Claimify} \cite{metropolitansky2025effectiveextractionevaluationfactual}. This framework evaluates generation through two complementary dimensions:
\begin{enumerate} \item \textbf{Exact Value Recall:} We extract specific data points—such as numbers and named entities—from the response and directly match them against the ground truth. This metric quantifies the model's ability to accurately retrieve and preserve atomic facts from the source.

\item \textbf{Semantic Claim Alignment:} We evaluate content quality using embedding-based precision and recall. This measures how well the response covers the reference information (Recall) and ensures the generated claims are factually grounded to minimize hallucinations.
\end{enumerate}

\section{Discussion and Limitations}\label{appendix:dal}

The effectiveness of our methodology stems from the construction of a semantic graph at the entry level, where each node represents a fine-grained table cell enriched with structural and contextual metadata. This granularity enables retrieval to directly align query semantics with individual cells, thereby bypassing the traditional necessity for Large Language Models (LLMs) to interpret complex raw layouts or infer latent hierarchical relationships from linearized text. Furthermore, the unified semantic space of the SAT graph inherently facilitates seamless cross-table retrieval, allowing for consistent reasoning across disparate data sources.

Despite these strengths, our approach is subject to several limitations. First, the framework's performance is highly contingent upon the quality of initial table parsing and metadata extraction. In real-world scenarios, tabular data often originates from unstructured formats like PDFs, necessitating Optical Character Recognition for conversion. Any errors in value identification or metadata extraction can propagate through the graph construction phase, ultimately degrading downstream retrieval accuracy. Second, our graph-based indexing faces scalability constraints. As the document corpus grows, the increased memory footprint and computational overhead associated with dense graph operations may pose challenges for large-scale industrial applications.

Nevertheless, these limitations present clear opportunities for future research. The integration of more robust, structure-aware parsing pipelines could significantly enhance the fidelity of the extracted graph. To address scalability, techniques such as graph sparsification, hierarchical indexing, or dynamic pruning offer promising avenues for optimizing the system's efficiency.

Furthermore, while the current work focuses on document embedded tables, we contend that the core philosophy of semantic units and graph-based representation is extensible to other structured data formats. This includes hierarchical tree-structured data such as XML and JSON, as well as complex spatial data structures. Adapting our framework to these diverse domains remains a direction for future investigation.
\begin{figure*}[t!] 
    \centering
    
    \begin{tcolorbox}[colback=gray!5, colframe=black!75, arc=1mm, 
    boxsep=0.5mm, left=1mm, right=1mm, top=0.5mm, bottom=0.5mm,
    title=\textbf{Prompt 1: Context-to-Entity Extraction (Metadata Restoration)}]
    
    \small
    \setlength{\parskip}{0.1em}
    
    \textbf{Role:} You are a Metadata Restoration Specialist. Your task is to EXTRACT the central entity name from an anonymized financial document where table headers lack subject identification.
    
    \textbf{Context:} The raw dataset suffers from anonymization where the central subject (e.g., company name or report topic) is often missing from table headers. You will receive the complete document content including title, introductory paragraphs, and table structures. Your goal is to identify and extract the precise entity name that serves as the global metadata field for cell-grouping.
    
    \textbf{Instructions:}
    \begin{enumerate}[leftmargin=*, nosep]
        \item \textbf{Entity Identification:} (a) Scan the document title and opening paragraphs for explicit entity mentions. (b) Analyze table content for implicit entity references (e.g., "our company"). (c) Identify the single entity that all tables and text describe.
        \item \textbf{Entity Types (Priority Order):} 
        \begin{itemize}[nosep]
            \item \textit{Company Name} (e.g., "Morgan Stanley"); \textit{Report Section} (e.g., "Note 8"); \textit{Financial Metric}; \textit{Event}.
        \end{itemize}
        \item \textbf{Strict Formatting Rules:} No markdown. No explanations. Extract the EXACT entity name as it appears.
    \end{enumerate}
    
    \textbf{Output Format:} Output ONLY a valid JSON object:
    \texttt{\{"entity": "<exact entity name>", "type": "<Company|Section|Metric|Event>"\}}
    
    \textbf{One-Shot Demonstration:}
    \textit{Input:} Document titled "Annual Report 2023"... mentions "The Company recorded..." and "Morgan Stanley's consolidated..."
    \textit{Output:} \texttt{\{"entity": "Morgan Stanley", "type": "Company"\}}
    \end{tcolorbox}
    
    \vspace{2mm} 
    
    \begin{tcolorbox}[colback=gray!5, colframe=black!75, arc=1mm, 
    boxsep=0.5mm, left=1mm, right=1mm, top=0.5mm, bottom=0.5mm,
    title=\textbf{Prompt 2: Stochastic QA Generation (Generate-or-Discard)}]
    
    \small
    \setlength{\parskip}{0.1em}
    
    \textbf{Role:} You are a Multi-Hop QA Validator. Your task is to generate a coherent question-answer pair from RANDOMLY PAIRED table cells, or REJECT the pairing if no logical relationship exists.
    
    \textbf{Context:} This is part of an Unbiased QA Generation pipeline using Stochastic Field Association. The cell pairs were selected via randomized shuffling and heuristic matching (exact string match on dates/headers), NOT semantic similarity. Your goal is to determine if a valid multi-hop question can be formed; if not, reject the pair.
    
    \textbf{Instructions:}
    \begin{enumerate}[leftmargin=*, nosep]
        \item \textbf{Feasibility Check (Generate-or-Discard):} (a) Analyze if the randomly paired cells share a logical relationship. (b) If the relationship is too tenuous or nonsensical, output REJECT. (c) Only proceed if a meaningful multi-dimensional question can be formed.
        \item \textbf{Question Generation (if feasible):} Create a question requiring cross-period, cross-metric, or cross-entity reasoning. Do NOT directly ask for specific values. Ensure practical business significance.
        \item \textbf{Answer Generation (if feasible):} Base strictly on provided data. Answer in 1-3 factual sentences. No predictions.
        \item \textbf{Context Integrity Constraints:} Each metric is ONLY valid within its complete context (entity + time + subject). Do NOT derive or compare across mismatched contexts.
    \end{enumerate}
    
    \textbf{Output Format:}
    \begin{itemize}[nosep]
        \item \textit{If valid:} \texttt{\{"question": "<multi-hop question>", "answer": "<data-grounded answer>"\}}
        \item \textit{If invalid:} \texttt{\{"reject": true, "reason": "<brief explanation>"\}}
    \end{itemize}
    
    \textbf{Data (Stochastically Paired):} \textit{\{context\}}
    \end{tcolorbox}
    
    \caption{The core prompt templates used in data preparation phase.}
    \label{fig:app_prompts}
\end{figure*}
\end{document}